\pdfoutput=1 

\documentclass[letterpaper,twocolumn,fleqn]{article} 

\usepackage{ist}
\usepackage{times}

\usepackage{soul}
\usepackage{url}
\usepackage[hidelinks]{hyperref}
\usepackage[utf8]{inputenc}
\usepackage[small]{caption}
\usepackage{graphicx}
\usepackage{amsmath}
\usepackage{booktabs}
\usepackage[symbol]{footmisc}

\usepackage{subfig}
\usepackage{xcolor}
\usepackage{hyperref}

\title{Towards Realistic Landmark-Guided Facial Video Inpainting
Based on GANs}

\author{Fatemeh Ghorbani Lohesara$^{(a)}$, Karen Eguiazarian$^{(b)}$, and Sebastian Knorr$^{(c)}$; \\$^{(a)}$ Communication Systems Group, Technische Universität Berlin, Berlin, Germany; \\$^{(b)}$ Computational Imaging Group, Tampere University, Tampere, Finland; \\$^{(c)}$ Ernst-Abbe University of Applied Sciences Jena, Jena, Germany;}

\date{} 

\begin{document} 

\maketitle

\thispagestyle{empty} 


\begin{abstract}
Facial video inpainting plays a crucial role in a wide range of applications, including but not limited to the removal of obstructions in video conferencing and telemedicine, enhancement of facial expression analysis, privacy protection, integration of graphical overlays, and virtual makeup. This domain presents serious challenges due to the intricate nature of facial features and the inherent human familiarity with faces, heightening the need for accurate and persuasive completions. In addressing challenges specifically related to occlusion removal in this context, our focus is on the progressive task of generating complete images from facial data covered by masks, ensuring both spatial and temporal coherence. Our study introduces a network designed for expression-based video inpainting, employing generative adversarial networks (GANs) to handle static and moving occlusions across all frames. By utilizing facial landmarks and an occlusion-free reference image, our model maintains the user's identity consistently across frames. We further enhance emotional preservation through a customized facial expression recognition (FER) loss function, ensuring detailed inpainted outputs. Our proposed framework exhibits proficiency in eliminating occlusions from facial videos in an adaptive form, whether appearing static or dynamic on the frames, while providing realistic and coherent results. 
\end{abstract}


\section{Introduction}
\label{sec:intro}

Inpainting, being an intricate task in computer vision, necessitates meeting critical criteria, including the meaningful integration of generated content with surrounding elements for semantic correctness and indistinguishable blending of filled-in regions. Image inpainting involves the process of contextually filling in missing regions within an image to ensure visual consistency. Video inpainting, on the other hand, extends the principles of image inpainting by introducing temporal constraints to ensure consistency across multiple frames. Within the realm of visual media containing occlusions, such as those arising from object removal, inpainting techniques strive to reconstruct missing content in a photorealistic and natural appearance. This challenge has recently drawn considerable attention due to the significant need for image and video editing applications across diverse industries. 

While extensive research has addressed image inpainting, video inpainting introduces additional complexities that remain largely unexplored. Moreover, existing studies have predominantly focused on scenarios involving object removal and scene inpainting, overlooking the distinct challenges posed by facial video inpainting, particularly when human subjects are involved. 

Facial video inpainting has diverse applications, ranging from occlusion removal in video conferencing and telemedicine to in-depth facial expression analysis, privacy preservation and identity verification systems, and enhancement of virtual makeup and beauty applications. For instance, privacy regulations mandate the non-release of patients' photo records without proper anonymization, often achieved through masking biometric information ~\cite{wu2020image, newton2005preserving}. This domain's difficulties arise from the complex nature of facial features and the inherent familiarity with faces, increasing the challenge of achieving a convincing completion. Moreover, current studies have primarily revolved around the removal of moving occlusions, denoted as moving masks, or static masks across frames ~\cite{chang2019free,chang2019learnable,zou2021progressive} as a separate problem. In broad terms, extending the idea of video inpainting to handle dynamic and static bounding box masks in videos has diverse applications. These include object tracking in video surveillance and autonomous driving systems, medical imaging, and graphic overlay in video content.

Our approach addresses these challenges by leveraging recent advancements in video inpainting, employing a generative adversarial network (GAN) to inpaint facial regions occluded by masks with different patterns of movements. Our adaptive pipeline takes inputs in the form of frames with applied masks, which can either be static or move across frames, a single reference frame without masks, and ground truth frames without occlusions. We detect facial landmarks from the latter set and incorporate them into our model alongside the masked frames and a reference frame.

Considering both dynamic and static masks across all frames, we conduct a comparative analysis of our framework with two existing models, LGTSM (our baseline model) and CombCN ~\cite{chang2019learnable, wang2019video}, demonstrating our model's superior performance in inpainting occlusions with varied types of movements. This evaluation employs a publicly available facial video dataset ~\cite{rossler2018faceforensics}.
The remaining sections of this paper are organized as follows: Firstly, we conduct a review of relevant literature on image and video inpainting methods. Afterward, we provide a detailed exposition of our proposed approach for facial video inpainting. We then proceed to present and analyze the experimental results, both qualitatively and quantitatively, based on the type of occlusion. The paper is ultimately concluded with a summary and a discussion of future directions for research.


\section{Related Work}
\label{sec:related_work}

\paragraph{Image Inpainting:} Image inpainting is the task of filling missing regions within an image in a visually consistent manner. Traditionally, this problem was approached using patch-based synthesis methods, as outlined in studies such as those by Efros et al. \cite{efros1999texture} and Barnes et al. \cite{barnes2009patchmatch}. While patch-based and diffusion-based approaches \cite{richard2001fast} demonstrated success in certain scenarios, they faced challenges, especially in dealing with complex structured images.

In recent years, the main focus in solving image inpainting problems has shifted to deep learning techniques, with significant advancements in the field of Generative Adversarial Networks (GANs) specially designed for image completion ~\cite{elharrouss2020image}. The dynamic domain of GANs has fulfilled the promise of generative models by producing realistic examples in various applications not limited to inpainting ~\cite{iglesias2023survey}. Notably, they have demonstrated advancement in image-to-image translation tasks, transforming photos, and generating photorealistic images that challenge human perception ~\cite{ko2023superstargan,de2023review}. 

GANs present an innovative approach to generative modeling, treating the problem as a learning task involving two sub-models: the generator, which generates new examples, and the discriminator, which classifies those samples as real or fake. The generator and discriminator work collaboratively to enhance image quality, resulting in images with heightened visual plausibility \cite{pathak2016context}. For example, an image inpainting network introduced by Iizuka et al. \cite{iizuka2017globally} incorporates discriminators operating at multiple scales, yet these approaches may require additional post-processing steps. In contrast, recent methodologies, such as the partial convolution method proposed by Liu et al. \cite{liu2018image}, offer post-processing-free alternatives to achieve similar outcomes.

\paragraph{Video Inpainting:} Video inpainting is essentially an extension of image inpainting, introducing temporal constraints to ensure coherence across various frames, as discussed in prior research ~\cite{yang2023deep, szeto2022devil, chang2019free, chang2019learnable, wu2020image, wang2019video}. Despite the extensive work in image inpainting, video inpainting presents its unique challenges that remain to be fully resolved. 

Notably, most existing studies have predominantly concentrated on scenarios involving object removal and scene inpainting ~\cite{zou2021progressive}, often overlooking the specialized realm of facial video inpainting involving human subjects. This area introduces additional complexities due to the intricate nature of facial features and the inherent familiarity of faces, rendering the task of achieving a convincing completion even more demanding. For video face inpainting, achieving temporal consistency is more critical. In this context, maintaining consistency in facial structures like eyes, and nose, as well as facial attributes such as facial hair, eyeglasses, and expressions should be considered carefully. Challenges specific to facial video inpainting can arise from occlusions caused by human-object interactions, dynamic backgrounds, clothing or accessories, and variations in lighting conditions. These complexities collectively hinder accurate facial feature analysis and reconstruction.

Moreover, the existing efforts in video inpainting, while promising, have primarily revolved around the removal of moving objects or individuals, denoted as moving masks, or static masks across frames. Moving masks results in the shifting of occluded regions' positions throughout the video sequence ~\cite{chang2019free,chang2019learnable,zou2021progressive}. However, in the scenarios where the masks are relatively big and consistent across the frames, the task of inpainting becomes more challenging, as the occluded area remains unchanged and there are no similar features in the neighbor frames for reconstructing the occluded frame correctly. 

While patch-based methods have excelled in video inpainting, they come with significant computational time constraints due to search algorithms. Furthermore, they face limitations in dealing with complex objects like faces. In ~\cite{wang2019video}, the authors proposed to jointly learn temporal-spatial structure for video inpainting, but masks are in a fixed shape and position across all frames, which does not hold true for face inpainting where the subject is in motion. Recently, a general video-to-video synthesis has been proposed ~\cite{wang2018video}; the proposed method utilizes optical flow information across frames to ensure temporal consistency and would require a large video dataset to ensure robustness to fine-grained face variations. 

Current facial video inpainting solutions do not effectively address both problems of static and moving mask removal, necessitating modifications to make them applicable in several real-world applications. Thus, there is a critical research gap for a new adaptive approach with the capability of inpainting both static and dynamic occlusions, specially designed for facial videos. 


\section{Architecture Overview}
\label{sec:arch}

Our framework's architectural design is centered around the Learnable Gated Temporal Shift Module (LGTSM), a model proposed by Chang et al. \cite{chang2019learnable} for video inpainting. The LGTSM optimizes 2D convolutions by intelligently shifting input channels to their temporal neighbors, enhancing temporal understanding crucial for video inpainting tasks. This design choice eliminates the need for additional parameters from 3D convolutions or optical flow data, resulting in a lightweight yet high-performance architecture. 

\begin{figure}[t]
  \centering
  \includegraphics[scale=0.4]{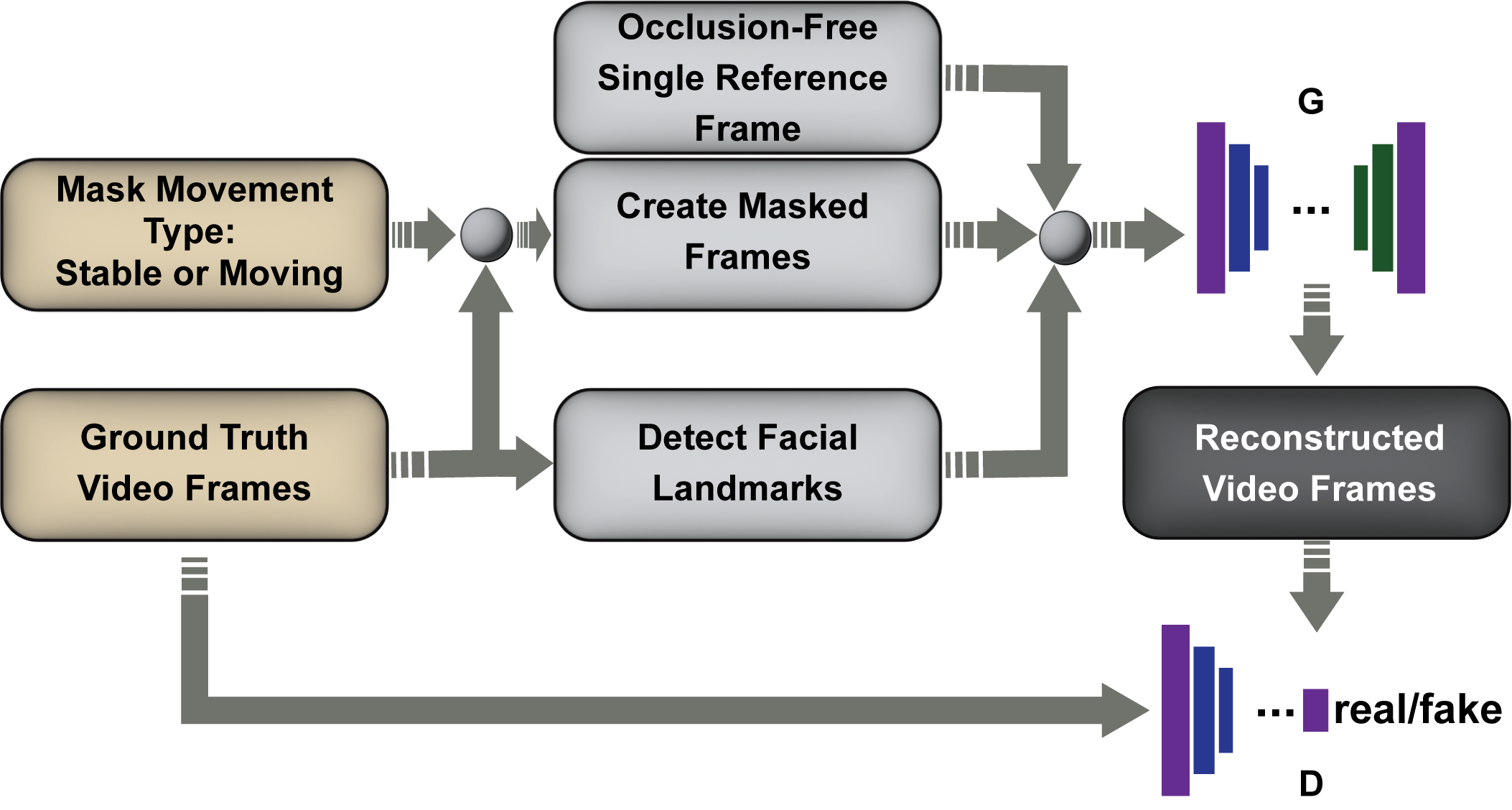}   
  \caption{Overview of the pipeline of the proposed GAN-based expression-aware inpainting with the support of facial landmarks and a single occlusion-free reference frame. The masked images and facial landmarks are provided as input to the generator (G) to synthesize the complete face images. The discriminator (D) then classifies generated faces as real or fake.
  }
    \label{fig:arch}
\end{figure}

To efficiently model temporal dynamics, we leverage the Temporal Shift Module (TSM) in its online form, as proposed by Lin et al. \cite{lin2019tsm}. This module enables temporal modeling by shifting the feature map along the temporal dimension without requiring future frame features suitable for real-time applications. Notably, TSM enhances temporal modeling capabilities at no additional computational cost on top of 2D convolutions.

To further help LGTSM in aggregating non-local information due to convolutional bias, our model is augmented with an attention mechanism \cite{zhang2019self}. This mechanism empowers the network to focus on diverse parts of the input data with regard to occlusion movement type e.g. moving or static, particularly improving global context understanding and capturing non-local features within the feature maps. 

It is worth noting that, the attention-driven, long-range dependency modeling facilitated by this mechanism plays an important role in image-generation tasks. Traditional convolutional GANs generate high-resolution details based solely on spatially local points in lower-resolution feature maps. By incorporating additional self-attention layers, our model can generate intricate details by considering cues from all feature locations. 

Our framework processes masked frames representing the occluded region considering their pattern of movement, a single reference frame without the mask, and facial landmarks as illustrated in Figure ~\ref{fig:arch}. The inclusion of an RGB reference face frame is essential for overcoming occlusion challenges and preserving the person's identity, ensuring accurate inpainting by considering individual facial features. For more details of the components of our proposed approach, we refer to ~\cite{lohesara2023EVIHRnet}, originally designed for the task of head-mounted display removal in Virtual Reality. This source provides a comprehensive exploration of the details underlying our method.

The generator in our model, comprising 13 convolution layers with the gated TSM, implements attention-based down-sampling, dilation, and up-sampling. Self-attention layers are strategically positioned to compute attention weights, allowing the network to capture spatial relationships, dependencies, and feature information within the input feature maps.

In the adversarial learning process, the discriminator evaluates inpainted frames against ground truth frames, compelling the generator to accurately fill occluded areas. This involves six 2D convolution layers with TSM, ensuring a comprehensive evaluation of the generated frames. The discriminator further ensures the consistency of highly detailed features across distant portions of the image, enhancing the overall visual fidelity.

Our model employs a combination of diverse loss functions for effective convergence. The L1 Reconstruction Loss emphasizes pixel-wise accuracy, measuring the fidelity of inpainted frames concerning ground truth frames. The VGG Loss based on ImageNet captures perceptual differences by utilizing a pre-trained VGG network on ImageNet, providing insights into high-level features ~\cite{simonyan2014very,russakovsky2015imagenet}. The Style Loss, inspired by Gatys et al. \cite{gatys2015neural}, ensures the preservation of stylistic features in the inpainted frames. The Wasserstein GAN Adversarial Loss further guides the generator to create realistic inpainted frames by fooling the discriminator ~\cite{arjovsky2017wasserstein}. The FER loss evaluates the model's performance in recognizing multiple facial expression classes designed based on ~\cite{savchenko2022video}, ensuring an accurate depiction of emotions in the inpainted frames, ultimately contributing to the overall accuracy of the model. 

These losses play a crucial role in guiding the training process, emphasizing factors such as reconstruction accuracy, perceptual differences, style variations, adversarial learning, and accurate replication of facial expressions, ensuring the generation of visually appealing and consistent facial video outputs.


\section{Experiments}
\label{sec:results}
In this section, we conduct a comprehensive comparison of our proposed method with other existing models in the literature for video inpainting, including our baseline model, LGTSM ~\cite{chang2019learnable}, and CombCN ~\cite{wang2019video}. CombCN, a two-stage deep video inpainting method, utilizes a 3D fully convolutional architecture for temporal structure inference and a 2D fully convolutional network for spatial detail recovery in image-based inpainting. We conduct these experiments employing both quantitative and qualitative assessments with random static and dynamic masks. 

For the implementation of our network, we leverage PyTorch version 1.10.0. In configuring the convolutional layers, we adopt a kernel size of $5\times5$ for the initial convolution layer, a $4\times4$ kernel size with a stride of 2 for down-sampling layers, and a $3\times3$ kernel size with dilation factors of 2, 4, 8, and 16 for the dilated layers. The remaining convolution layers utilize a $3\times3$ kernel size. The attention layers employ a $1\times1$ kernel size. The activation function employed throughout is the LeakyReLU. For optimization during training, we utilize the Adam optimizer with a learning rate set to $9.8\times 10^{-5}$.

Finally, the weights assigned to the overall loss function in our model, are set as 1, 4, 10, 1, and 1 for Adversarial, FER, Style, VGG, and L1 Reconstruction losses, respectively. These weights play an important role in emphasizing the contribution of each loss component to the overall optimization objective during the training process.

Given the limited availability of facial video datasets compared to image datasets suitable for learning-based models, we employ the FaceForensics ~\cite{rossler2018faceforensics} dataset in this study. This dataset comprises 1,004 videos with over 500,000 frames featuring faces of newscasters collected from YouTube, with most videos containing frontal faces cropped to a size of $128\times128$ pixels—ideal for training purposes. For testing, we use 150 videos with a duration of 32 frames, while the remaining videos contribute to training the models. All models undergo training on the FaceForensics dataset using random static and dynamic bounding boxes as outlined in \cite{chang2019free} to ensure a fair comparison.

\subsection{Quantitative results}
Quantitatively, we evaluate the models using various metrics, including mean square error (MSE), peak-signal-to-noise ratio (PSNR), and structural similarity index (SSIM) to assess image quality. It is noteworthy that these metrics provide detailed insights into the quality of the inpainting results. Additionally, we report Learned Perceptual Image Patch Similarity (LPIPS) and Fréchet inception distance (FID) score as evaluation metrics, known for their alignment with human judgments of image similarities.

The first assessment involves applying static masks to the video frames, and the comparative evaluation is presented in Table \ref{tab:metrics_static}. Our proposed model demonstrates superior performance compared to CombCN and LGTSM across a spectrum of evaluation metrics.  As outlined in Table \ref{tab:metrics_static}, our proposed model excels by achieving the lowest MSE, LPIPS, and FID scores, signifying minimized errors and perceptual discrepancies. Moreover, it attains the highest PSNR and SSIM values, indicating better quality and structural fidelity in the context of static occlusion removal.


Shifting the focus to the task of moving mask removal, our proposed model maintains competitive performance, surpassing CombCN and LGTSM across the evaluation metrics summarized in Table \ref{tab:metrics}. The same set of metrics is employed to provide a comprehensive assessment of the inpainting quality in scenarios involving moving masks. 
As evident in the table, our model achieves the lowest MSE, showcasing a 36.36\% improvement over LGTSM and a notable 58.82\% improvement over CombCN. The model also excels in LPIPS, where it outperforms LGTSM by 31.69\% and CombCN by an impressive 71.05\%. Additionally, our model exhibits superior FID scores, indicating a 14.22\% improvement over LGTSM and a substantial 37.91\% improvement over CombCN.
Remarkably, our model attains the highest values in PSNR, boasting a 5.85\% increase over LGTSM and a remarkable 13.80\% increase over CombCN. Similarly, in SSIM, our model outshines with a 1.04\% increase over LGTSM and a notable 3.01\% increase over CombCN.
It is noteworthy that, despite the integration of online TSM in LGTSM, our model consistently outperforms LGTSM in real-time scenarios. This observation underscores the efficacy of our proposed method in handling dynamic occlusions in real-world settings.
Furthermore, in our investigation into the impact of online and offline TSM usage in both our model and LGTSM, where TSM is an integral part of the network, we conducted an ablation study. This study aims to elucidate the influence of leveraging temporal information from future neighboring frames in the context of moving mask removal. The inclusion of offline TSM proves particularly advantageous when masks are in motion. This feature enables both models to leverage information from future frames without occlusion in the inpainting of the current frame's masked features, resulting in outputs of higher quality. As indicated in Table \ref{tab:metrics}, our model exhibits the highest performance when employing offline TSM compared to the offline counterpart in LGTSM.

\subsection{Qualitative results}
\begin{figure}[!th]
  \centering
  \includegraphics[width=0.6\columnwidth]{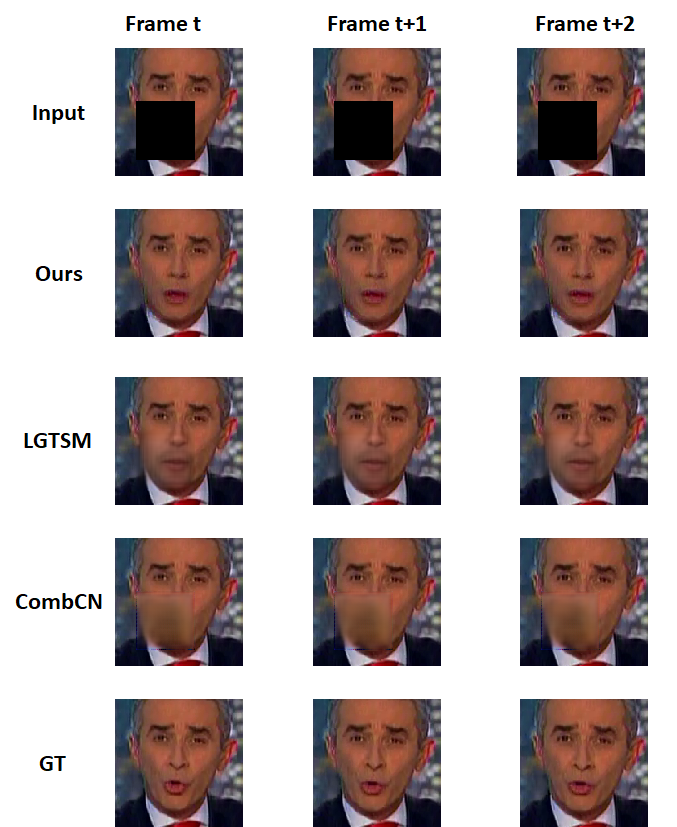}
  \caption{Sample of inpainted frames in FaceForensics validation set (ID 18) resulted from our model, LGTSM, and CombCN, along with the corresponding input and GT frames. The applied masks are static on the frames. Images: RCN TV (\url{https://www.youtube.com/watch?v=8ILvKPA3TI0})}
  \label{fig:comparison2}
\end{figure}

In qualitative evaluations, our model demonstrates significant improvements when handling static masks, as highlighted in Figure ~\ref{fig:comparison2}. The inpainted frames from our model exhibit a remarkably closer resemblance to the ground truth (GT) frames compared to the other models. This visual evidence underscores the efficacy of our model in generating realistic outputs, showcasing its proficiency in addressing occlusion removal and preserving facial structure and expressions. This success is attributed to strategic elements, including the utilization of a single reference image in the masked area, the integration of FER loss, and the incorporation of facial landmarks ~\cite{lohesara2023EVIHRnet}.

\begin{figure}[t]
  \centering
  \includegraphics[width=0.6\columnwidth]{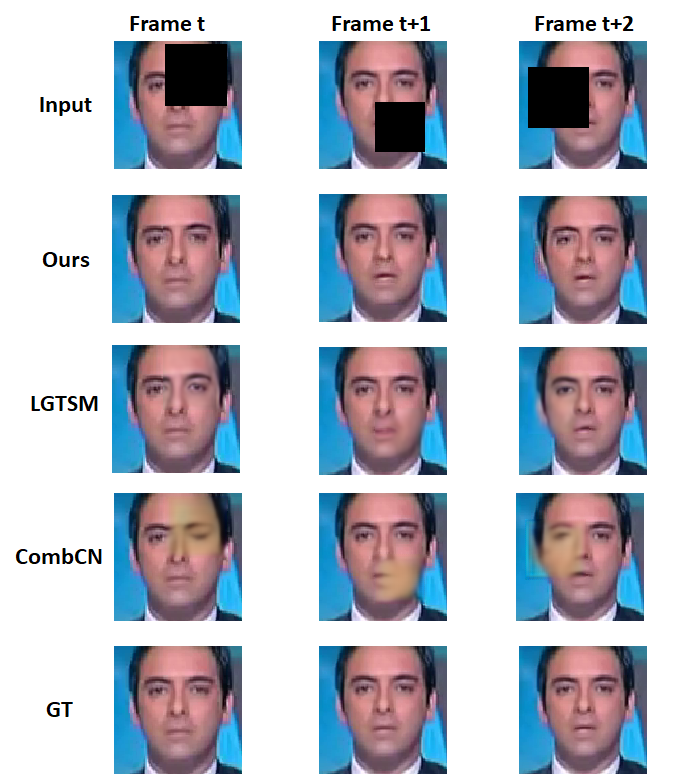}
  \caption{{Sample of inpainted frames in FaceForensics validation set (ID 73) resulted from our model, LGTSM, and CombCN, along with the corresponding input and GT frames. The masks vary along the frames. Images: MTV Lebanon News (\url{https://www.youtube.com/watch?v=irbGBNQaZ1E})}}
  \label{fig:comparison}
\end{figure}

Similarly, in scenarios involving moving masks, as illustrated in Figure ~\ref{fig:comparison}, our model indicates a more satisfactory similarity between the inpainted frames and the GT frames compared to alternative models. This competitive performance compared to our baseline can be attributed to several key factors: attention usage challenges, the limited necessity for reference usage, and the adoption of an online inpainting strategy in contrast to the offline approach. In scenarios involving moving masks, the usage of reference images may not be inherently necessary for this specific task as the model can recover and learn the occluded features from the neighbor frames and also future frames in offline learning. Furthermore, When occlusions occur in diverse areas across different frames, the model encounters challenges in capturing long-range dependencies through its attention mechanism. In contrast, in stable mask scenarios, our model reveals significant performance which is attributed to the beneficial reference usage and attention mechanism. These elements play a pivotal role in ensuring accurate and visually pleasing inpainting results in such a situation.

In summary, our model consistently outperforms LGTSM and CombCN, highlighting its effectiveness in recovering occluded areas with both moving and static masks. This superiority is particularly evident when leveraging offline TSM, where features from future frames contribute to learning, or when solely addressing static masks across entire frames. Notably, the enhanced performance is more evident in scenarios involving static masks, where our model excels in maintaining accurate facial shapes, such as lips, overcoming challenges observed in other models as can be observed in ~\ref{fig:comparison2}.

Additional visual comparisons and videos with respect to the diversity of the subjects and reference frames used for inpainting can be found in the supplementary materials. 
\begin{table}[!h]
\caption{Quantitative results of FaceForensics validation set with static masks. The metrics are averaged resulted from our model, the baseline model (LGTSM), and the CombCN model.}
\label{tab:metrics_static}
\begin{center}       
\begin{tabular}{|p{1cm}|p{1cm}|p{2cm}|p{2cm}|} 
\hline
\textbf{Model} & \textbf{Ours} & \textbf{LGTSM} \textbf{\cite{chang2019learnable}} & \textbf{CombCN} \textbf{\cite{wang2019video}}\\
     \hline
    \textbf{MSE\(\downarrow\)} & \textbf{0.0013} & 0.0017 & 0.0022 \\ \hline
    \textbf{PSNR\(\uparrow\)} & \textbf{30.01} & 28.45 & 27.27 \\ \hline
    \textbf{SSIM\(\uparrow\)} & \textbf{0.9525} & 0.9418 & 0.9354 \\ \hline  
    \textbf{LPIPS\(\downarrow\)} & \textbf{0.0317} & 0.0437 & 0.0831 \\ \hline 
    \textbf{FID\(\downarrow\)} & \textbf{0.5974} & 0.6626 & 0.7973  \\ \hline
\end{tabular}
\end{center}
\end{table} 

\begin{table*}[!h]
\caption{Quantitative results of FaceForensics validation set with moving masks. The metrics are averaged resulted from our online model, the baseline model (online LGTSM), and the CombCN model. The results of our model and LGTSM with offline TSM are also described. }
\label{tab:metrics}
\begin{center}       
\begin{tabular}{|p{1cm}|p{2.5cm}|p{2.5cm}|p{3cm}|p{3cm}|p{1.5cm}|}
\hline
\textbf{Model} & \textbf{Ours with Offline TSM} & \textbf{Ours with Online TSM} & \textbf{LGTSM with Offline TSM} \textbf{\cite{chang2019learnable}} & \textbf{LGTSM with Online TSM} \textbf{\cite{chang2019learnable}} &\textbf{CombCN} \textbf{\cite{wang2019video}}\\ 
     \hline
    \textbf{MSE\(\downarrow\)} & \textbf{0.0006} & 0.0007 & 0.0009 & 0.0011 & 0.0017 \\ \hline
    \textbf{PSNR\(\uparrow\)} & \textbf{32.67} & 32.05 & 30.98 & 30.27 & 28.17 \\ \hline
    \textbf{SSIM\(\uparrow\)} & \textbf{0.9662} & 0.9615 & 0.9592 & 0.9516 & 0.9334 \\ \hline  
    \textbf{LPIPS\(\downarrow\)} & \textbf{0.0254} & 0.0304 & 0.0354 & 0.0446 & 0.1078 \\ \hline 
    \textbf{FID\(\downarrow\)} & \textbf{0.5762} & 0.6629 & 0.6703  & 0.773 & 1.067 \\ \hline
\end{tabular}
\end{center}
\end{table*} 

\section{Conclusion}
\label{sec:conclusion}
Facial video inpainting emerges as a key research problem with widespread applications, ranging from video conferencing and medical imaging by eliminating occlusions to enhancing facial expression analysis, improving security systems, and refining virtual makeup. This field presents specific challenges, necessitating solutions that can deliver realistic and convincing completions.
Our expression-based video inpainting network, anchored in generative adversarial networks (GANs), adaptively addresses challenges posed by both static and moving occlusions. By intelligently leveraging facial landmarks and an unoccluded reference image, our model smoothly preserves the user's identity across frames. The incorporation of a FER loss function further helps emotional preservation, yielding outputs that are not only realistic but also emotionally detailed. Beyond maintaining facial expressions and identity coherently across frames, our model exhibits temporal consistency throughout the inpainted sequences. 

Future work in video inpainting holds promising directions, notably in the domains of higher-resolution 2D video inpainting and 3D volumetric video inpainting. For higher-resolution 2D video inpainting, the focus lies in refining neural architectures to accommodate increased data complexity, ensuring the preservation of intricate facial details at high resolutions. This advancement is vital for applications such as high-quality video conferencing and medical imaging. Simultaneously, delving into 3D volumetric video inpainting opens avenues for immersive virtual and augmented reality experiences. Adapting neural networks to handle the temporal and spatial intricacies of 3D video data will be key for practical deployment in such real-world scenarios.

\bibliographystyle{IEEEbib}
\bibliography{paper}

\begin{biography}
Fatemeh Ghorbani Lohesara is currently a Marie Curie Fellow and a Ph.D. student in the Communication Systems Group at Technische Universität Berlin, Germany. She received her M.Sc. degree in mechatronics from K. N. Toosi University, and holds a B.Sc. in electrical engineering from Guilan University. Her research focuses on addressing the headset removal problem for gaze contact in XR applications. Her main research interests include human–computer interaction, XR, and serious games.
\newline

Karen Eguiazarian (Fellow,
IEEE) (SM’96, F’18) received the M.Sc. degree in mathematics from Yerevan State University, Armenia, in 1981, the Ph.D. degree in physics and mathematics from Moscow State University, Russia, in 1986, the Doctor of Technology degree from the Tampere University of Technology (TUT), Tampere, Finland, in 1994. He is a Professor with the Signal Processing Department, Tampere University, leading the Computational Imaging Group, and a Docent with the Department of Information Technology, University of Jyväskylä, Finland. His main research interests are in the fields of computational imaging, compressed sensing, efficient signal processing algorithms, image/video restoration, and image compression. He has published over 750 refereed journal and conference articles, books, and patents in these fields. Prof. Egiazarian is a member of the DSP Technical Committee of the IEEE Circuits and Systems Society. He has served as an associate editor in major journals in the field of his expertise, including the IEEE Transactions on Image Processing, and was Editor-in-chief of the Journal of Electronic Imaging (SPIE).
\newline

Sebastian Knorr (SM'19, IEEE) received the Dipl.-Eng. and Dr.-Eng. degree in electrical engineering from Technical University of Berlin in 2002 and 2008, respectively. He is professor at the Ernst Abbe University of Applied Sciences Jena, leading the Innovation Center for Immersive Imaging Technologies (3IT Jena). His main research interests are in the field of computer vision, 3D image processing and immersive media, in particular virtual reality applications. 
Prof. Knorr received the German Multimedia Business Award of the Federal Ministry of Economics and Technology in 2008, and was awarded by the initiative “Germany-Land of Ideas” which is sponsored by the German government, commerce and industry in 2009, respectively. He received a couple of best paper awards including the Scott Helt Memorial Award of the IEEE Transactions on Broadcasting in 2011 and the Lumiére Award at the International Conference on 3D Immersion in 2018. Prof. Knorr has served as an associate editor for the IEEE Trans. on Multimedia and is currently serving as an associate editor for the IEEE Trans. on Image Processing. 
\newline

\section{Acknowledgments}
\small
This project has received funding from the European Union’s Horizon 2020 research and innovation program under the Marie Skłodowska-Curie grant agreement No 956770.

\end{biography}

\end{document}